\documentclass[conference]{IEEEtran}
\IEEEoverridecommandlockouts
\usepackage{cite}
\usepackage{amsmath,amssymb,amsfonts}
\usepackage{algpseudocode}
\usepackage{algorithm}
\usepackage{algorithmicx}
\usepackage{graphicx}
\usepackage{textcomp}
\usepackage{xcolor}
\usepackage[T5,T1]{fontenc}
\newcommand{\rowvec}[1]{[\begin{matrix}#1\end{matrix}]}
\newcommand{\abs}[1]{\left | #1\right | }
\newcommand{\norm}[1]{\left\lVert#1\right\rVert}
\newtheorem{comment}{Comment}
\newtheorem{lemma}{Lemma}

\begin{document}

\title{A Solution for a Fundamental Problem of 3D Inference based on 2D Representations}

\author{\IEEEauthorblockN{\fontencoding{T5}\selectfont Thi\d\ecircumflex{}n \Acircumflex{}n L. Nguy\~\ecircumflex{}n}
\IEEEauthorblockA{\textit{thienannguyen.cv@gmail.com}}
}

\maketitle
\thispagestyle{plain}
\pagestyle{plain}

\begin{abstract}
3D inference from monocular vision using neural networks is an important research area of computer vision. Applications of the research area are various with many proposed solutions and have shown remarkable performance. Although many efforts have been invested, there are still unanswered questions, some of which are fundamental. In this paper, I discuss a problem that I hope will come to be known as \textit{a generalization of the Blind Perspective-n-Point (Blind PnP) problem for object-driven 3D inference based on 2D representations}. The vital difference between the fundamental problem and the Blind PnP problem is that 3D inference parameters in the fundamental problem are attached directly to 3D points and the camera concept will be represented through the sharing of the parameters of these points. By providing an explainable and robust gradient-decent solution based on 2D representations for an important special case of the problem, the paper opens up a new approach for using available information-based learning methods to solve problems related to 3D object pose estimation from 2D images.
\end{abstract}


\section{Introduction}
Learning properties of 3D objects from 2D images is a typical module of 3D inference from monocular vision and is also a common task in computer vision applications. Based on the success of 2D convolutional neural networks in understanding semantic information of 2D images \cite{9356353}, these neural networks were quickly adapted to extract 3D information from 2D images \cite{wang2018pixel2mesh}, \cite{DBLP:journals/corr/abs-1904-07850}. Among approaches of using 2D convolutional neural networks for extracting 3D information from 2D images, differentiable rendering approaches are diverse and have many advantages in improving performance \cite{DBLP:journals/corr/abs-2006-12057}, \cite{c1e01e12a6174337977a562567d13a35}. A key point of differentiable rendering techniques is a differentiable rendering pipeline that allows flowing gradients from rendered pixels to parameters of the neural network. However, the variety of graphical representations make differentiation of rendering process is not uniquely defined \cite{DBLP:journals/corr/abs-2006-12057}. Inspired by regular grids of convolutional layers, as well as the desire for making contributions in this paper benefit as many differentiable rendering techniques as possible, I introduce a novel graphical representation based on the result of translating a certain graphics code \cite{DBLP:journals/corr/KulkarniWKT15} to a grid of pixel-level 3D information under the support of a linear perspective projection, called \textit{2D representation of 3D information}. 

Instead of constructing a complete solution of extracting 3D information from 2D images, I focus on a fundamental problem of inverse rendering. Thus, I will construct a dataset that represents an important special case of the problem, propose a solution for the special case and analyze the solution on the dataset. The dataset, Sky\footnote{https://github.com/thienannguyen-cv/Sky-dataset}, was constructed on the ApolloCar3D dataset \cite{Song2019ApolloCar3DAL} but was modified to balance practicality and complexity to make it simple enough for analysis. Because of the nature of the ApolloCar3D dataset, lights and material appearance are not taken into account. However, they are not the purpose of this paper. 

In summary, my contributions are: 
\begin{itemize}
\item Describe a fundamental problem that can be applied for differentiable rendering techniques. 
\item Introduce a dataset that represents an important special case of the problem. 
\item Propose a solution for the special case, demonstrate and analyze the performance of the solution on the dataset. 
\end{itemize}

\section{A Fundamental Problem}
\subsection{2D Representation}
Define a \textit{fragment} $g$ is a tuple of coordinates and rendering information of a 3D point has form $g=(coord(g),info(g))=((x,y,z) \in \mathbb{R}^{3},r \in \mathbb{R}^{+})$, where $coord(\cdot)$ and $info(\cdot)$ are, respectively, functions permit accessing the coordinates and the rendering information value of a \textit{fragment}. Given a set of $K$ \textit{fragments} $S=\{g_{i}\}^{K}_{i=1}$, a \textit{2D representation} of $S$ on a screen has size $H \times W$ using a linear perspective projection (without extrinsic parameters) $P:\mathbb{R}^{3} \rightarrow \mathbb{R}^{2}$ is a function $L^{H \times W}_{S,P,AGG}:[1..H] \times [1..W]  \rightarrow \mathbb{R}$, 
\begin{equation}
\setlength{\nulldelimiterspace}{0pt}
	L^{H \times W}_{S,P,AGG}(i,j)=\left\{\begin{IEEEeqnarraybox}[\relax][c]{l's}
		AGG(T_{S,P}(i,j)), & if $T_{S,P}(i,j) \ne \emptyset$, \\
		0, & otherwise, %
	\end{IEEEeqnarraybox}\right.
\end{equation}
where $AGG(\cdot)$ is an aggregation function of \textit{fragments}' rendering information, and $T_{S,P}(i,j)$ returns a set of \textit{fragments} in $S$ which have $(i,j)$ as the result of applying the projection $P$ to their coordinates ($(i,j)$ is called screen position of \textit{fragments} in the set). Precisely, $T_{S,P}(i,j)=\{g \in S|\lfloor P(coord(g)) \rfloor =(i,j)\}$ in that the floor function $\lfloor \cdot \rfloor$ was expanded for tuples, 
\begin{equation}
\lfloor (x_{1}, x_{2}, \dots, x_{n}) \rfloor \triangleq (\lfloor x_{1} \rfloor, \lfloor x_{2} \rfloor, \dots, \lfloor x_{n} \rfloor). 
\end{equation}

For the simpleization of notations, the matrix notation $L_{ij}$ usually used to substitute $L^{H \times W}_{S,P,AGG}(i,j)$. 

\textbf{Proper 2D representation. } Considering a function $f:\mathbb{R}^{+} \rightarrow \mathbb{R}^{3}$ that decodes coordinates of a 3D point from a value of rendering information. We say a \textit{2D representation} $L$ is \textit{proper} under the function $f$ if $(i,j)=\lfloor P(f(L_{ij})) \rfloor$, for every $(i,j)$ satisfies $L_{ij} \ne 0$. 

\subsection{An aggregation function}
Consider a set of $N$ \textit{fragments} $S=\{((x_{i},y_{i},z_{i}),r_{i})\}^{N}_{i=1}$. Let $z_{min}=min(\{z_{i}|z_{i}>0\}^{N}_{i=1})$ is the minimum depth value of visible points of $S$, define $rmin(\cdot)$ aggregation function: 
\begin{equation}
rmin(S) \triangleq min(\{r|((x,y,z),r) \in S_{z_{min}}\}), 
\end{equation}
where $S_{z_{min}}=\{((x,y,z),r) \in S|z=z_{min}\}$. Thus, $rmin(S)$ returns the minimum render information value of closest-to-camera visible points of $S$. A convention that $rmin(\emptyset)=0$. 

\subsection{Problem Description}
Given a set of $N$ 3D points $Q=\{x_{i} \in \mathbb{R}^{3}\}^{N}_{i=1}$, a screen has size $H \times W$ and a linear perspective projection (without extrinsic parameters) $P:\mathbb{R}^{3} \rightarrow \mathbb{R}^{2}$, $Q'=\{x'_{i}=(x^{(1)}_{i}+a,x^{(2)}_{i}+b,x^{(3)}_{i}+c)|x_{i} \in Q\}^{N}_{i=1}$ is a set of $N$ 3D points generated by adding unknown amounts $a$, $b$, and $c$ to coordinates of each point in $Q$. Let $F'=\{((x'^{(1)}_{i},x'^{(2)}_{i},x'^{(3)}_{i}),1)|x'_{i} \in Q'\}^{N}_{i=1}$ is a set of \textit{fragments} constructed on $Q'$. Let $Q$, $P$ and the value of the \textit{2D representation} $L^{H \times W}_{F',P,rmin}$ is known, find the values of $a$, $b$, and $c$? 

\section{Sky Dataset}
\begin{figure}[htbp]
\centerline{\includegraphics{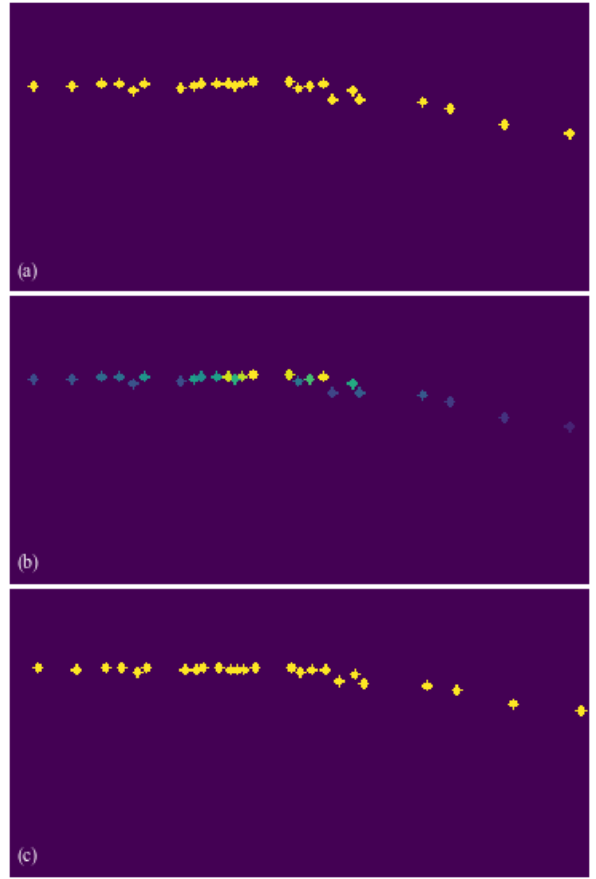}}
\caption{A configuration in the Sky dataset. (a) the binary mask of the \textit{2D representation} $E'$. (b) the \textit{2D representation} $E'$ (brighter is bigger). (c) the target \textit{2D representation} $L$ of the configuration. }
\label{fig1}
\end{figure}

The dataset is constructed on the ApolloCar3D dataset, a dataset of urban street views with car-object 3D pose annotations. Fig.~\ref{fig1} is a sample in the Sky dataset. While the ApolloCar3D dataset contains 5,277 driving images of $3384 \times 2710$ resolution and over 60K car instances of 79 car models \cite{Song2019ApolloCar3DAL}, the Sky dataset is built only upon 3D annotations of their 4,248 images which are chosen randomly. With $K$ is the projection was used in the ApolloCar3D dataset, construct configurations of the problem (names of variables are still remain) for each image in 4,248 chosen images: 
\begin{itemize}
\item Let $O=\{c_{i}=(x_{i},y_{i},z_{i})\}^{M}_{i=1}$ is the set of locations of car objects in the 3D annotation of the image, $G = \{((x_{i},y_{i},z_{i}) \in O,z_{i})\}^{M}_{i=1}$ is a set of \textit{fragments} constructed on $O$. 
\item Let $P(x) = K(x) \odot \rowvec{\frac{256}{2710} & \frac{256}{3384}}^{\intercal}  - \rowvec{128 & 0}^{\intercal}$ is a projection constructed on the projection $K$. Let $E^{128 \times 256}_{G,P,rmin}$ is a \textit{2D representation} of $G$ on a screen has size $128 \times 256$. 
\item A \textit{2D representation} $E'^{128 \times 256}_{F,P,rmin}$ is constructed on $E$ by drawing uniform circles (described in the Appendix) at non-zero points (active points) in $E$. Thus, $Q$ is the set of $N$ 3D points are reconstructed from $E'$ so that projection of a 3D point in $Q$ is also screen position of an active point on $E'$. Also, a point in $E'$ corresponds to only one point in $Q$, ignored obscured points. 
\item According to the fundamental problem, I construct a \textit{2D representation} $L^{128 \times 256}_{F',P,rmin}$ from $Q$, $P$ ($L$ is called \textit{target 2D representation} of the configuration). With $a=0.517$, $b=0.303$ and $c=0$,  $F'=\{((x'^{(1)}_{i},x'^{(2)}_{i},x'^{(3)}_{i}),1)|x'_{i} \in Q'\}^{N}_{i=1}$ is a set of $N$ \textit{fragments} created from $Q'$, where $Q'=\{x'_{i}=(x^{(1)}_{i}+a,x^{(2)}_{i}+b,x^{(3)}_{i})|x_{i} \in Q\}^{N}_{i=1}$. 
\end{itemize}
In summary, the dataset is 4,248 configurations of a special case of the fundamental problem where $c$ is ignored, the projection $P$, the unknowns $a$ and $b$ is unique. 

\section{A Gradient Descent Solution}
Consider the special case of the problem with its 4,248 instances in the Sky dataset. According to the description of the dataset, we have a rendering algorithm $R$ that takes a set of $N$ 3D points of a configuration, $Q$, and returns an image $A$ that is also a \textit{2D representation} $E'^{128 \times 256}_{F,P,rmin}=A=R(Q)$. In that, $P$ is the projection in the Sky dataset. And, $F=\{(x_{i}=(x^{(1)}_{i},x^{(2)}_{i},x^{(3)}_{i}),r)|x_{i} \in Q, rinv(r_{i})=(x^{(1)}_{i},x^{(2)}_{i},x^{(3)}_{i})\}^{N}_{i=1}$, where $rinv$ is a differentiable function. Via the function $rinv$, we imply that spatial information in values of a point in $E'$ is linked to its screen position on $E'$. In other words, $A$ is a \textit{proper 2D representation} under the function $rinv$ (\textit{proper 2D representation} for short). Let $\theta^{*}=(a,b)$ is the ground truth parameter and $\widetilde{\theta}=(\widetilde{a}, \widetilde{b})$ is its estimation, $\widetilde{I}$ is an image constructed on $A$, 
\begin{equation}
\setlength{\nulldelimiterspace}{0pt}
	\widetilde{I}_{ij}=\left\{\begin{IEEEeqnarraybox}[\relax][c]{l's}
		add(A_{ij},\widetilde{\theta}_{3D}), & if $A_{ij} \ne 0$, \\
		0, & otherwise, %
	\end{IEEEeqnarraybox}\right.
\end{equation}
where $\widetilde{\theta}_{3D}=(\widetilde{a},\widetilde{b},0)$ is represented $\widetilde{\theta}$ in 3D space and the differentiable function $add(\cdot)$ adds $\widetilde{\theta}_{3D}$ to the coordinates of 3D points encoded in the rendering information $A_{ij}$, 
\begin{equation}
rinv(\widetilde{r}=add(r,\widetilde{\theta}_{3D})) \triangleq rinv(r)+(\widetilde{a},\widetilde{b},0). 
\end{equation}

As a part of the solution, I propose a differentiable mapping algorithm which permits the gradient propagates to $\widetilde{\theta}$, called \textit{Spatial Domain Mapping} (SDM), for generating a \textit{proper 2D representation} $\widetilde{L}$ from $\widetilde{I}$. After the forward pass of \textit{SDM}, spatial information of a point in $\widetilde{I}$ is linked to its screen position (called \textit{proper screen position}, at this point). So, we have 
\begin{equation}
\widetilde{L}^{128 \times 256}_{\widetilde{F},P,rmin}(i,j)=rmin(q(\widetilde{I},(i,j))),
\label{eq:wL}
\end{equation}
where $q(\widetilde{I},(i,j))$ returns a set of \textit{fragments} constructed from 3D points encoded in $\widetilde{I}$ which have $(i,j)$ as the result of applying the projection $P$ on them, 
\begin{equation}
q(\widetilde{I},(i,j))=\{(rinv(\widetilde{I}_{kl}),\widetilde{I}_{kl})|(i,j)=\lfloor P(rinv(\widetilde{I}_{kl})) \rfloor \}_{k,l}. 
\label{eq:q}
\end{equation}

Note that, for a configuration in the Sky dataset, the value of the target \textit{2D representation} $L$ of the configuration does not encode spatial information, however, it is a binary mask. This does not allow learning the parameter $\widetilde{\theta}$ directly through matching between $\widetilde{L}$ and $L$. Therefore, I need to construct a loss function $J$ of two \textit{2D representations} are $\widetilde{L}$ and $L$, so that the solution can be considered as an optimization problem: 
\begin{equation}
\theta^{*} = \underset {\widetilde{\theta}} {\text{arg min }} J(\widetilde{L},L). \label{eq:opt}
\end{equation}
I use the gradient descent method \cite{Ruder2016AnOO} and a technique called \textit{dynamic gradient flow} to solve the optimization problem. The main idea behind \textit{dynamic gradient flow} is considering matching the binary mask of $\widetilde{L}$ and $L$ as a multi-objective optimization problem where each point corresponds to an objective. Thus, for each objective, I control the gradient flow in a way that helps to translate an existing point to its corresponding screen position instead of ``bubbling" a point at a certain screen position, what is a common process found in solutions based on image loss optimization. The gradient flow will be mentioned in the backward pass of \textit{SDM}. 

\subsection{Spatial Domain Mapping}
For the convenience of explanation, I define functions $xinv(\cdot)$, $yinv(\cdot)$, and $zinv(\cdot)$ through the below equation: 
\begin{equation}
rinv(r) \triangleq (xinv(r), yinv(r), zinv(r)). 
\end{equation}

\subsubsection{Forward pass}
\begin{figure}[htbp]
\centerline{\includegraphics{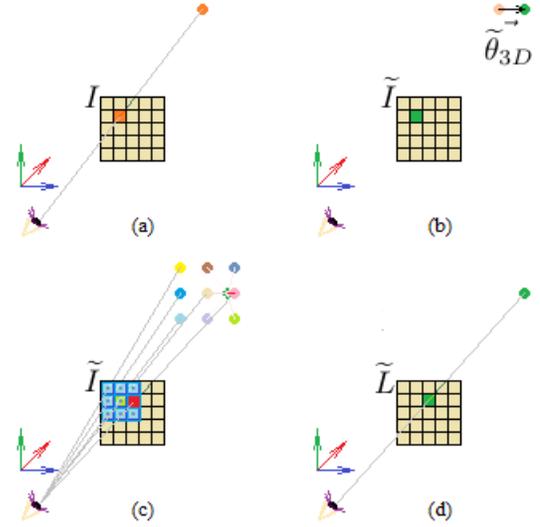}}
\caption{Illustration of the forward pass of the SDM algorithm. The kernel $D_{22}$ is indicated by a cyan $3 \times 3$ grid in (c). Zero elements in kernels in $D$ are expressed by transparent (red) squares. (a) an active point (the orange square) in $I$ and its corresponding 3D point (the orange circle). (b) the input of the algorithm, the \textit{2D representation} $\widetilde{I}$, is the result of adding $\widetilde{\theta}$ to the active point. (c) values of elements in $D_{22}$ are exponentially inversely-proportional to distances (darker is smaller) from the 3D point represented by $\widetilde{I}_{22}$ (the dark green circle) to its corresponding \textit{XY-adjustments} (other circles). In this case, $D_{2223}$ (the opaque red square) is the maximum element of $D_{22}$. (d) the output of the algorithm. }
\label{fig2}
\end{figure}

The forward pass of the algorithm generates the \textit{proper 2D representation} $\widetilde{L}$ from the parameter $\widetilde{\theta}$ in a differentiable way. This is illustrated in Fig.~\ref{fig2}. Thus, for each point $\widetilde{I}_{ij}$, define a kernel $D_{ij}$ has size $h \times w$ ($h$, $w$ are odds) to indicate the offset of its \textit{proper screen position} relative to its current screen position by a real value in the interval $[0,1]$. Let $(c_{h} \triangleq \lceil h/2 \rceil,c_{w} \triangleq \lceil w/2 \rceil)$ is the position of the center element of the kernel. 
If $\widetilde{I}_{ij} \ne 0$, each point $D_{ijkl}$ in the kernel corresponds to a 3D point $T_{ijkl}=(x_{ijkl},y_{ijkl},z_{ijkl})$ has the same depth value as $rinv(\widetilde{I}_{ij})$ and its screen position via applying the linear perspective projection $P$ is $(i+k-c_{h}+0.5,j+l-c_{w}+0.5)$ ($T_{ijkl}$ is called the \textit{XY-adjustment} of $\widetilde{I}_{ij}$ at the screen position $(i+k-c_{h},j+l-c_{w})$), 
\begin{IEEEeqnarray}{c}
\text{For }k \in [1..c_{h}],l \in [1..c_{w}]: z_{ijkl}=zinv(\widetilde{I}_{ij}), \label{eq:Tz}\\
(i+k-c_{h}+0.5,j+l-c_{w}+0.5)= P(T_{ijkl}). \label{eq:TP}
\end{IEEEeqnarray}
Otherwise, $\widetilde{I}_{ij}=0$, I set a safe value for \textit{XY-adjustments}, $T_{ijkl}=(0,0,0)$. Precisely, $D_{ij}$ is the result of applying a \textit{softmax}-ish function to distances from the 3D point represented by $\widetilde{I}_{ij}$ to its \textit{XY-adjustments}, $\{T_{ijkl}\}_{k,l}$, 
\begin{equation}
\setlength{\nulldelimiterspace}{0pt}
	D_{ijkl} \triangleq \left\{\begin{IEEEeqnarraybox}[\relax][c]{l's}
		\frac{\exp(c \norm{rinv(\widetilde{I}_{ij})-T_{ijkl}})}{\sum\limits_{k,l}\exp(c \norm{rinv(\widetilde{I}_{ij})-T_{ijkl}})}, & if $\widetilde{I}_{ij} \ne 0$, \\
		\frac{1}{hw}, & otherwise, %
	\end{IEEEeqnarraybox}\right.
\label{eq:D}
\end{equation}
where, in cases that $I_{ij} \ne 0$, $c$ is a shape constant for controlling the difference between the max value, which corresponds to the nearest \textit{XY-adjustment}\footnote{Ties are ignored as they are rare boundary conditions. While, the backward pass in the general case described in this paper is not affected by them. }, and other values of $D_{ij}$. Here, I assume that $c$ is large enough so, approximately, the max value of $D_{ij}$ is $1$ and thus, other values are $0$. \\

\begin{comment}
Let $B_{ijkl}=D_{ijkl} \cdot \widetilde{I}_{ij}$, 
\begin{equation}
\begin{split}
C_{ij} \triangleq &rmin(\{(rinv(B_{(i-k+c_{h})(j-l+c_{w})kl}), \\
&B_{(i-k+c_{h})(j-l+c_{w})kl})\}_{1 \le k \le c_{h}, 1 \le l \le c_{w}})
\end{split}
\label{eq:C}
\end{equation}
is a $H \times W$ matrix. We have $\lim\limits_{c \to \infty} C_{ij} = \widetilde{L}^{128 \times 256}_{\widetilde{F},P,rmin}(i,j)$. 
\label{com:1}\\
\end{comment} 

Proofs of all Comments and Lemmas can be found in the Appendix. Finally, through applying the result of \textit{Comment~\ref{com:1}}, the matrix $C$ is the output of the forward pass. 

\subsubsection{Backward pass}
The backward pass refers to the gradient flow from the loss function $J$ to the parameter $\widetilde{\theta}$ in computing the gradient $\nabla_{\widetilde{\theta}} J(\widetilde{L},L)$. This is illustrated in Fig.~\ref{fig3}. To understand the solution, let's consider a basic case and then the general case: 

\begin{figure*}[tb]
\center{\includegraphics[width=\textwidth]{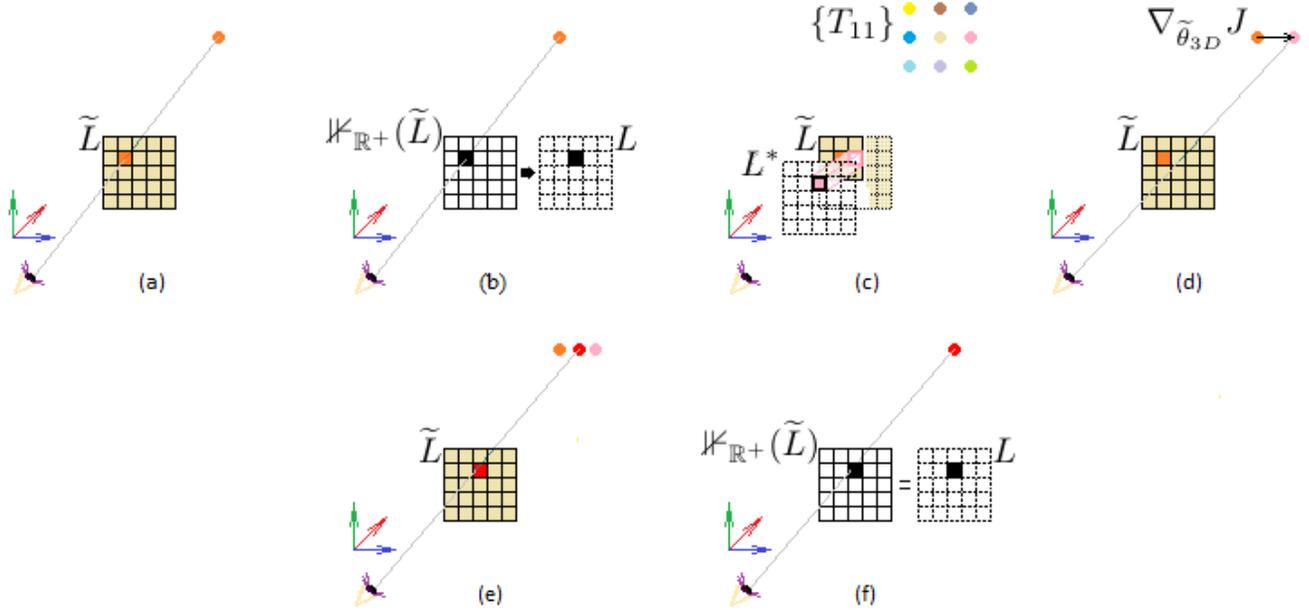}}
\caption{Illustration of the backward pass of the SDM algorithm in a single-point example. The active point in $\widetilde{I}$ is $\widetilde{I}_{({u}_{0}=2)({v}_{0}=2)}$. (a) the output of the SDM algorithm, the \textit{proper 2D representation} $\widetilde{L}$. (b) the binary mask of $\widetilde{L}$ and the target \textit{2D representation} $L$ (dash lines). (c) the brown solid grid indicates points in $\widetilde{L}$ which depend on $\widetilde{I}_{22}$ so that the gradients from these points will be passed to $\widetilde{I}_{22}$. The \textit{proper 2D representation} $L^{*}$ is constructed on $L$ and the \textit{XY-adjustment} $T_{2223}$. The projection of the active point at the screen position $(u'_{0},v'_{0})=(2,3)$ in $L^{*}$ on $\widetilde{L}$ shows the point which has gradient in $\widetilde{L}$. (d) the gradient $\nabla_{\widetilde{\theta}} J(\widetilde{L},L)$ is projected on the 3D space (the black vector). (e) the optimal \textit{proper 2D representation}. (f) the binary mask of optimal \textit{proper 2D representation} and the target \textit{2D representation}. }
\label{fig3}
\end{figure*}
\textbf{Single-point. } Assume the set $Q$ has only one element. Consider the corresponding point of that element in $\widetilde{I}$, $\widetilde{I}_{{u}_{0} {v}_{0}} \ne 0$, and its kernel $D_{{u}_{0} {v}_{0}}$. Determine $\widetilde{k}_{0}$ and $\widetilde{l}_{0}$ so that the max value of $D_{{u}_{0} {v}_{0}}$ is $D_{{u}_{0} {v}_{0} \widetilde{k}_{0} \widetilde{l}_{0}} \approx 1$. Let $\Upsilon_{0}$ is a set of positions of $N'_{0}$ points in $\widetilde{L}$ which depend on $\widetilde{I}_{{u}_{0} {v}_{0}}$, $\Upsilon_{0} \triangleq \{(\mu^{i}_{0},\nu^{i}_{0})|{u}_{0} - c_{h} < \mu^{i}_{0} < {u}_{0} + c_{h}, {v}_{0} - c_{w} < \nu^{i}_{0} < {v}_{0} + c_{w}\}^{N'_{0}}_{i=1}$. As the gradient from these points will be passed to $\widetilde{I}_{{u}_{0} {v}_{0}}$, defining a loss function based on these points will help with optimizing $\widetilde{\theta}$. Under the assumption of a single point, I create a \textit{proper 2D representation} from the binary mask $L$ by assigning to the active point of $L$ the property of $\widetilde{I}_{{u}_{0} {v}_{0}}$. Let $L_{u'_{0} v'_{0}}$ be the target active point of $\widetilde{I}_{{u}_{0} {v}_{0}}$, define a \textit{proper 2D representation} $L^{*}$, 
\begin{equation}
\setlength{\nulldelimiterspace}{0pt}
	L^{*}_{ij} \triangleq \left\{\begin{IEEEeqnarraybox}[\relax][c]{l's}
		r_{{u}_{0} {v}_{0} (i-{u}_{0}+c_{h}) (j-{v}_{0}+c_{w})}, & if $(i,j)=(u'_{0},v'_{0})$, \\
		0, & otherwise, %
	\end{IEEEeqnarraybox}\right.
\label{eq:Lstar}
\end{equation}
where $rinv(r_{ijkl}) = T_{ijkl}$. I assume that $(u'_{0},v'_{0}) \in \Upsilon_{0}$. \\

\begin{comment}
Let 
\begin{equation}
\theta_{opt} \triangleq \underset {\widetilde{\theta}} {\text{arg min }} \sum\limits_{i,j}\abs{rinv(\lim\limits_{c \to \infty} \widetilde{L}_{ij})-rinv(L^{*}_{ij})}. 
\end{equation}
If $\widetilde{\theta} = \theta_{opt}$ then the binary mask of $\widetilde{L}$ is equal to $L$ ($\mathbb{1}_{\mathbb{R}^{+}}(\widetilde{L})=L$). 
\label{com:2}\\
\end{comment}

\noindent Apply this result to \eqref{eq:opt}, we have 
\begin{equation}
J(\widetilde{L},L) = \sum\limits_{i,j}\abs{rinv(\widetilde{L}_{ij})-rinv(L^{*}_{ij})}
\end{equation}

\noindent as the suitable loss function. 

Consider points in $\widetilde{L}$ and $L^{*}$ which have screen positions in $\Upsilon_{0}$, as only these points can affect the value of the gradient flow at $\widetilde{I}_{{u}_{0} {v}_{0}}$. I realize that, according to \eqref{eq:C}, there will be at most two active points are $\widetilde{L}_{({u}_{0}+\widetilde{k}_{0}-c_{h})({v}_{0}+\widetilde{l}_{0}-c_{w})}$ and $L^{*}_{u'_{0} v'_{0}}$, as shown in Fig.~\ref{fig3}. Thus, let $(\widetilde{u}_{0},\widetilde{v}_{0}) \triangleq ({u}_{0}+\widetilde{k}_{0}-c_{h},{v}_{0}+\widetilde{l}_{0}-c_{w})$ and by the chain rule, we have 
\begin{IEEEeqnarray}{rCl}
\frac{\mathrm{d} J}{\mathrm{d} \widetilde{\theta}} &=& \frac{\mathrm{d} J}{\mathrm{d} \widetilde{I}_{{u}_{0} {v}_{0}}}\frac{\mathrm{d} \widetilde{I}_{{u}_{0} {v}_{0}}}{\mathrm{d} \widetilde{\theta}}, \label{eq:J} \\ 
&=& \frac{\mathrm{d} \abs{rinv(\widetilde{L}_{\widetilde{u}_{0} \widetilde{v}_{0}})-rinv(L^{*}_{\widetilde{u}_{0} \widetilde{v}_{0}})}}{\mathrm{d} \widetilde{I}_{{u}_{0} {v}_{0}}}\frac{\mathrm{d} \widetilde{I}_{{u}_{0} {v}_{0}}}{\mathrm{d} \widetilde{\theta}} \nonumber \\
&&+\> \frac{\mathrm{d} \abs{rinv(\widetilde{L}_{u'_{0} v'_{0}})-rinv(L^{*}_{u'_{0} v'_{0}})}}{\mathrm{d} \widetilde{I}_{{u}_{0} {v}_{0}}}\frac{\mathrm{d} \widetilde{I}_{{u}_{0} {v}_{0}}}{\mathrm{d} \widetilde{\theta}}, \label{eq:JL}
\end{IEEEeqnarray}
when $(\widetilde{u}_{0},\widetilde{v}_{0}) \ne (u'_{0},v'_{0})$. In case $(\widetilde{u}_{0},\widetilde{v}_{0}) = (u'_{0},v'_{0})$, the first term is decayed. However, my purpose is matching the binary mask of $\widetilde{L}$ and $L$ so, if nothing is specified, we imply that 
\begin{equation}
(\widetilde{u}_{0},\widetilde{v}_{0}) \ne (u'_{0},v'_{0}). \label{eq:ignore}
\end{equation}
In this case, because of the assumption of a single point, we have $L^{*}_{\widetilde{u}_{0} \widetilde{v}_{0}} = 0$. Thus, the first term of \eqref{eq:JL} supports moving the \textit{proper screen position} of $\widetilde{I}_{{u}_{0} {v}_{0}}$ away the $(\widetilde{u}_{0},\widetilde{v}_{0})$ position. And so, the second term supports moving it toward the $(u'_{0},v'_{0})$ position. Now, I can simplify the gradient harmlessly by canceling the first term and removing the role of the kernel $D$ in computing the gradient: 
\begin{equation}
\frac{\mathrm{d} J}{\mathrm{d} \widetilde{\theta}} = \frac{\mathrm{d} \abs{rinv(\widetilde{I}_{u_{0} v_{0}})-rinv(r_{{u}_{0} {v}_{0} {\Delta' u}_{0} {\Delta' v}_{0}})}}{\mathrm{d} \widetilde{I}_{{u}_{0} {v}_{0}}}\frac{\mathrm{d} \widetilde{I}_{{u}_{0} {v}_{0}}}{\mathrm{d} \widetilde{\theta}}, \label{eq:Je}
\end{equation}
where $({\Delta' u}_{0}, {\Delta' v}_{0}) \triangleq (u'_{0}-{u}_{0}+c_{h},v'_{0}-{v}_{0}+c_{w})$. This is equivalent to reducing the distance between two 3D points encoded in the rendering information of $\widetilde{I}_{u_{0} v_{0}}$ and of $L^{*}_{u'_{0} v'_{0}}$. 

\textbf{Multiple points. } This case brings us back to the problem described in the Sky dataset. Let $W=\{(u_{i},v_{i})|\widetilde{I}_{u_{i} v_{i}} \ne 0\}^{M'}_{i=1}$ is the set of positions of $M'$ active points in $\widetilde{I}$. Thus, to be able to apply the same solution as the single-point case for each element in $W$, these two problems must be solved: 
\begin{itemize}
\item \eqref{eq:JL} is only valid for the single-point case. 
\item Let $U=\{J_{i}\}^{M'}_{i=1}$ is the set of loss functions for active points in $\widetilde{I}$. How will these loss functions be aggregated? 
\end{itemize}

Consider a set $\{\Upsilon_{i}\}^{M'}_{i=1}$, $\Upsilon_{i}$ is constructed from $(u_{i},v_{i})$ in the same way as constructing the $\Upsilon_{0}$ from $(u_{0},v_{0})$ in the single-point case. From $\{\Upsilon_{i}\}^{M'}_{i=1}$, construct a set $\{\Upsilon'_{i}\}^{M'}_{i=1}$. Thus, for each element $(u_{i},v_{i})$ in $W$, $\Upsilon'_{i}$ is a set of screen positions of $N_{i}$ active points in $\widetilde{L}$ which depend on $\widetilde{I}_{u_{i} v_{i}}$ except the \textit{proper screen position} $(\widetilde{u}_{i},\widetilde{v}_{i})$ of $\widetilde{I}_{u_{i} v_{i}}$ on $\widetilde{L}$, 
\begin{equation}
\Upsilon'_{i} \triangleq \{(u'^{(j)}_{i},v'^{(j)}_{i}) \in \Upsilon_{i} \setminus \{(\widetilde{u}_{i},\widetilde{v}_{i})\}|L_{u'^{(j)}_{i} v'^{(j)}_{i}} \ne 0\}^{N_{i}}_{j=1}. \label{eq:grad_flow}
\end{equation}
In that, $(u'^{0}_{i},v'^{0}_{i})$ is always the target active points of $(u_{i},v_{i})$. So, the valid expansion for the gradient at \eqref{eq:J} is
\begin{IEEEeqnarray}{rCl}
\frac{\mathrm{d} J_{i}}{\mathrm{d} \widetilde{\theta}}&=&\frac{\sum\limits_{j}\mathrm{d} \abs{rinv(\widetilde{I}_{{u}_{i} {v}_{i}})-rinv(L^{*}_{u'^{(j)}_{i} v'^{(j)}_{i}})}}{\mathrm{d} \widetilde{I}_{{u}_{i} {v}_{i}}}\frac{\mathrm{d} \widetilde{I}_{{u}_{i} {v}_{i}}}{\mathrm{d} \widetilde{\theta}} \nonumber \\
&&+\> \frac{\mathrm{d} \abs{rinv(\widetilde{I}_{{u}_{i} {v}_{i}})-rinv(L^{*}_{u'^{0}_{i} v'^{0}_{i}})}}{\mathrm{d} \widetilde{I}_{{u}_{i} {v}_{i}}}\frac{\mathrm{d} \widetilde{I}_{{u}_{i} {v}_{i}}}{\mathrm{d} \widetilde{\theta}}. \label{eq:Ji}
\end{IEEEeqnarray}

For the second problem, because loss functions in $U$ are independent of each other, all loss functions can be optimized by optimizing their summation 
\begin{equation}
\mathcal{J}(\widetilde{L},L) \triangleq \sum\limits_{i} J_{i}(\widetilde{L},L). \label{eq:ls}
\end{equation}

\subsection{Optimization and Convergence}
In a sense, the Sky dataset was dividing a problem into small configurations. This helps with preventing the binary \textit{2D representation} $L$ from being full of active points, which means it doesn't contain any useful infomation for estimation. On the other hand, as a result, an optimization method for the problem must be able to estimate $\theta^{*}$ through configurations. This makes the stochastic gradient descent (SGD) \cite{Ruder2016AnOO} is the most natural and suitable choice for optimization. \\

\begin{comment}
Assume that  active points in the binary mask $L$ are distributed independently, $H$ and $W$ are much bigger than $h$ and $w$, respectively, and the size of the kernels (and the size of the screen) are big enough to be considered, in the scope of this comment, as boundless. We have: 
\begin{itemize}
\item The first terms of \eqref{eq:Ji} are independent random variables. 
\item The expectation of the first term is zero. 
\item The variance of the first term is less than a constant. 
\\
\end{itemize}
\label{com:3}
\end{comment}

The assumption of distributing points independently in \textit{Comment~\ref{com:3}} is only reasonable for the positions of circles’ centers. However, through zooming \textit{2D representations} out as will be mentioned in the \textit{implementation details} section, we can ignore the shape of circles. This helps retain the result of \textit{Comment~\ref{com:3}} for the Sky dataset. \\

\begin{lemma}
Consider a set of 3D points $Q=\{q_{i} \in \mathbb{R}^{3}\}^{N}_{i=1}$, where $N>0$ is the number of elements of $Q$ and a tuple $\theta=(a,b)$. Let $\widetilde{\theta}=(\widetilde{a},\widetilde{b})$ be a tuple of unknowns. $Q'=\{x'_{i}=(q^{(1)}_{i}+a,q^{(2)}_{i}+b,q^{(3)}_{i})|q_{i} \in Q\}^{N}_{i=1}$ and $\widetilde{Q}=\{\widetilde{x}_{i}=(q^{(1)}_{i}+\widetilde{a},q^{(2)}_{i}+\widetilde{b},q^{(3)}_{i})|q_{i} \in Q\}^{M}_{i=1}$ ($0 < M \le N$) are constructed by adding $\theta$ and $\widetilde{\theta}$ to the coordinates of points in $Q$. Let $F'=\{(x_{i},r_{i})|x_{i} \in Q',r_{i} \in \mathbb{R}^{+}\}^{N}_{i=1}$, and $\widetilde{F}=\{(\widetilde{x}_{i},\widetilde{r}_{i})|\widetilde{x}_{i} \in \widetilde{Q},\widetilde{r}_{i} \in \mathbb{R}^{+}\}^{M}_{i=1}$ be two sets of \textit{fragments} constructed on $Q'$ and $\widetilde{Q}$. Let $L'^{H \times W}_{F',P,rmin}$ and $\widetilde{L}^{H \times W}_{\widetilde{F},P,rmin}$ be \textit{2D representations} constructed on $F'$ and $\widetilde{F}$, where $P$ is a linear perspective projection (without extrinsic parameters). Assume that there exists points $p \in Q$, $p' \in Q'$, $\widetilde{p} \in \widetilde{Q}$, and a screen position $(u,v)$. So that, $p=(X,Y,z)$, $p'=(X+a,Y+b,z)$, and $\widetilde{p}=(X+\widetilde{a},Y+\widetilde{b},z)$, where $X$ and $Y$ are independent random variables. And, $(u,v)$ is not the projection on $\widetilde{L}$ of any 3D point in $\widetilde{Q}$ other than $\widetilde{p}$. Assume the probability that $(u,v)$ is the projection on $L'$ of $p'$ is $c$ in $(0,1) \subset \mathbb{Q}$. Thus, if $\widetilde{L}$ always has the same binary mask as $\widetilde{L} \odot L'$, 
\begin{equation}
P(\mathbb{1}_{\mathbb{R}^{+}}(\widetilde{L})=\mathbb{1}_{\mathbb{R}^{+}}(\widetilde{L} \odot L'))=1, 
\label{eq:Pr}
\end{equation}
then $\lim\limits_{z \to 0} \widetilde{\theta}=\theta$. 
\label{lem:1}\\
\end{lemma}

Although the result of \textit{Lemma~\ref{lem:1}} is for ideal data, there is a more practical result that can apply to almost all kinds of data. Thus, if the screen has an ideal density (unlimited fined) the result can be applied for any $z$ (without the $\lim$ notation). 

Known that the SGD algorithm accumulates gradients calculated from independent samples to update the parameter $\widetilde{\theta}$, if the learning rate $r$ is small enough then the variance of the contribution of the first term of \eqref{eq:Ji} in the optimization will decay to zero. Thus, based on \textit{Comment~\ref{com:3}}, the loss function was defined in \eqref{eq:ls} is still suitable for the optimization problem. Based on \textit{Comment~\ref{com:2}} and \textit{Lemma~\ref{lem:1}}, I construct a convergence criterion. Thus, I estimate the probability in \eqref{eq:Pr} through supervising the empirical expectation over $n$ consecutive times with $n$ is large enough for applying the law of large numbers in the estimation, and stop the SGD algorithm if it is equal to $0$. While the empirical convergence is still assured, the learning rate $r$ will be decided empirically through hyperparameter tuning \cite{Hutter2019-ld}. 

\section{Experiments}
I split randomly 768 configurations of the Sky dataset, which were chosen randomly, into two datasets are train set and dev set. Thus, the train set has 512 configurations for training and the dev set has 256 configurations for validation. 

\subsection{Implementation Details}
I used the Pytorch deep learning framework for implementation. There are many ways for implementing the algorithm but my criteria was utilizing built-in functions of the framework and simplifying implementation. I chose the size of each mini-batch as large as possible and it was $16$. The convergence criterion was substituted by using the \textit{number of epochs} concept \cite{Afaq2020SignificanceOE}, \cite{Ruder2016AnOO}. Thus, the algorithm will pass over the entire dataset many times until it is almost sure that the algorithm was converged. The number of epochs $n_{epoch}$ and the learning rate $r$ were determined through hyperparameter tuning, as shown in Table~\ref{tab1}. 

Additionally, instead of tuning the hyperparameters $h$, $w$, I chose $(h, w) = (3, 3)$ and used a training algorithm as a valid substitution for expanding the kernels’ size. Precisely, I optimized the parameter $\widetilde{\theta}$ with smaller \textit{2D representations} through an algorithm that reduces the size of \textit{2D representations} by half in all dimensions and generates four new \textit{zoomed-out 2D representations}. Because the distance between two points in 3D space is directly proportional to their distance in the projection plane, I propose a divide-and-conquer algorithm for reshaping. Thus, a point in a new \textit{2D representation} is mapped into a set of points in the original \textit{2D representation} which are nearest to the center point of these points on the original screen. And, to be still able to use the result of \textit{Comment~\ref{com:2}}, screen positions used to construct \textit{XY-adjustments} must reflect the positions of these center points on the original screen. 

For each training batch, the size of the \textit{2D representation} $\widetilde{I}$ is reduced by $2^{s}$ in all two dimensions to generate $4^{s}$ new \textit{2D representations}, where $s$ is a positive integer. This can be done by applying the reshape algorithm multiple times to \textit{2D representations}. Then, the \textit{SDM} algorithm is performed iteratively along with increasing the size of \textit{2D representations} until these \textit{2D representations} reach the desired size. Precisely, after performing the \textit{SDM} algorithm, the size of \textit{2D representations} is increased (doubled in all dimensions) by inverting the reshape algorithm. This implementation helps with a valid forward pass as the original \textit{SDM} algorithm. However, it makes the backward pass different from the original algorithm (now, the \textit{2D representation} $\widetilde{I}$ for the backward pass is the output of the forward pass instead of its original version). 
\begin{table}[htbp]
\caption{The Performance of the Algorithm among Hyperparameter Configurations}
\begin{center}
\begin{tabular}{|c|c|c|}
\hline
\multicolumn{1}{|c|}{\textbf{Loss}}&\multicolumn{2}{|c|}{\textbf{Hyperparameters}} \\
\cline{2-3} 
\multicolumn{1}{|c|}{\textbf{value}} & \textbf{\textit{number of epochs ($n_{epoch}$)}}& \textbf{\textit{learning rate ($r$)}} \\
\hline
$6.703$& $1$& $1e-4$ \\
\hline
$\mathbf{0.930}^{*}$& $\mathbf{1}$& $\mathbf{3e-4}$ \\
\hline
$1.004$& $1$& $1e-3$ \\
\hline
$1.355$& $2$& $1e-4$ \\
\hline
$0.957$& $2$& $3e-4$ \\
\hline
$1.227$& $2$& $1e-3$ \\
\hline
\multicolumn{3}{l}{$^{*}$ The best performance.}
\end{tabular}
\label{tab1}
\end{center}
\end{table}
Nonetheless, this difference is needed to guarantee the results of \textit{Comment~\ref{com:3}} for $3 \times 3$ kernels. 

A zoomed-out target \textit{2D representation} is used for optimization. The idea is a correct zoom-out version of the target \textit{2D representation} will help the algorithm still intact the results of \textit{Comment~\ref{com:2}} and \textit{Lemma~\ref{lem:1}}. Finally, after convergence of the algorithm for zoomed-out \textit{2D representations}, the size of these \textit{2D representations} is increased (again, by inverting the reshape algorithm) for optimizing the loss function with nearer target active points until convergence of the algorithm for the original size. Additionally, the learning rate $r$ should be reduced by half as the distances between \textit{XY-adjustments} of a point are decreased along with the increase of the size of the \textit{2D representation}. 

The pseudocode of the SDM algorithm can be found in the Appendix.

\subsection{Evaluation Method}
I evaluate the algorithm on criteria are accuracy and robustness. 

\subsubsection{Accuracy}
Instead of creating a test set for considering the performance of the algorithm, I evaluate directly based on the Euclidean distance between the parameter $\theta^{*}$ and its estimation, $\widetilde{\theta}$. 

\subsubsection{Robustness}
\begin{figure*}[tb]
\center{\includegraphics[width=\textwidth]{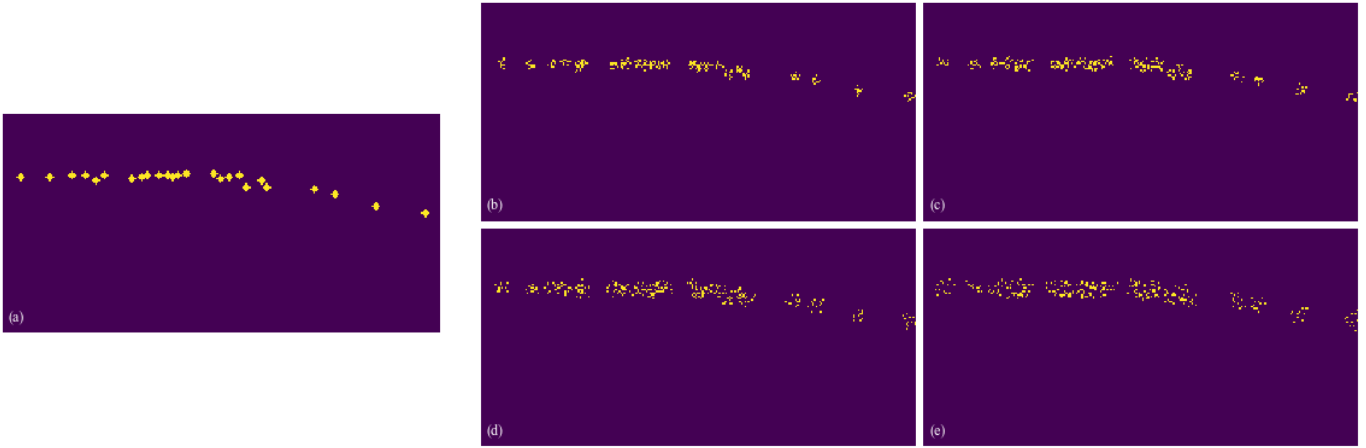}}
\caption{Effect of the noise on the target \textit{2D representation} of a sample in the Sky dataset. (a) the original target \textit{2D representation}. (b), (c), (d), and (e) the corresponding target \textit{2D representations} at \textit{$1$-pixels}, \textit{$2$-pixels}, \textit{$3$-pixels}, and \textit{$4$-pixels} noise levels. }
\label{fig4}
\end{figure*}

To evaluate the robustness, I add noise into the train set to understand the behavior of the algorithm against noise, which is a common characteristic of real data especially in computer vision. The noise will be added to the target \textit{2D representations}. Precisely, considering a non-negative integer $n$, I randomly change screen positions of active points in the target \textit{2D representations}. Thus, for each target \textit{2D representation} $L$, let $\{(u_{i}, v_{i})|L_{u_{i} v_{i}} \ne 0\}^{N}_{i=1}$ is the set of screen positions of active points in $L$, where $N$ is the number of active points. I construct a new target \textit{2D representation} $L'$ to substitute for $L$, so that the set of screen positions of active points in the new target \textit{2D representation} is $\{(u_{i}+X_{i}, v_{i}+Y_{i})\}^{N}_{i=1}$, where $X_{1},  \dots, X_{N}, Y_{1}, \dots, Y_{N}$ are identically distributed independent random variables uniformly distributed in $[-n..n]$. So that, $n$ is called \textit{noise level}. I also use the term \textit{$n$-pixels} to mention $n$ because it affects screen positions. The effect of the noise on the data is illustrated in Fig.~\ref{fig4}. 

\subsection{Result Analysis}
I chose $s=4$ which helps the algorithm considers transitions at a range of up to $16$ pixels on the screen. Via hyperparameter tuning, I decided $n_{epoch}=1$ and $r=3e-4$ (the best performance configuration in Table~\ref{tab1}). I considered noise levels are \textit{0-pixels} for evaluating the accuracy criteria and \textit{$1$-pixel}, \textit{$2$-pixels}, \textit{$3$-pixels}, and \textit{$4$-pixels} for evaluating the robustness criteria. For each noise level, I performed the experiment $60$ times. The result is summarized in Table~\ref{tab2}. 

For comparison, I exchange the noise levels to Euclidean distance (the second column in Table~\ref{tab2}) through interpolating the transition of XY coordinates of a point based on its screen position and Z coordinate. The first row of Table~\ref{tab2} describes the accuracy of the algorithm in the noise-free environment is $0.000682 \pm 0.00068404$, with a $95\%$ confidence level. Compared with the average deviation ($0.2834$) in the case of the smallest level of noise (\textit{$1$-pixel} noise-level), the accuracy is usually less than one hundred times smaller than the average deviation. As also shown in Table~\ref{tab2}, the average performance of the algorithm ($0.040056$) even in the worst case of noise (\textit{$4$-pixels} noise-level) is much less than the average deviation in the case of the \textit{$1$-pixel} noise-level. Thus, $75\%$ of outcomes are less than a sixth of the average deviation. Also, the value of the mean plus triple the standard deviation is less than a fourth of the average deviation. Although distributions are slightly right-skewed, the evidence is still strong enough to demonstrate that the algorithm is accurate and stable against noise. 

\section{Conclusion}
The problem mentioned in this paper is basic and directly related to more complex and practical problems in 3D inference from synthesis 2D views. Despite the number of parameters is small (only two parameters), compared to the size of the train set, being there are no explicit links between a 3D point and its corresponding representation on synthesis 2D views is the real challenge in solving the problem. Fortunately, with some practical assumptions, I could come up with an optimization solution for a major special case of the problem. The solution is stable against noise and has many potentials for further improvements in the future. 
\begin{table}[htbp]
\caption{Statistics of the Euclidean Distance between the Parameter $\theta^{*}$ and Its Estimations for All Experiments}
\begin{center}
\begin{tabular}{|c|c|c|c|c|c|c|}
\hline
\textbf{Noise} & \textbf{Avg. Dist.} & \multicolumn{5}{|c|}{\textbf{Statistics quantities ($1e-3$)}} \\
\cline{3-7} 
\textbf{level} & ($1e-3$) & \textbf{\textit{mean}} & \textbf{\textit{std}} & \textbf{\textit{Q1}} & \textbf{\textit{Q2}} & \textbf{\textit{Q3}} \\
\hline
0 & 0. & 0.682 & 0.349 & 0.410 & 0.677 & 0.915 \\
\hline
1 & 283.4 & 16.340 & 3.528 & 13.525 & 15.623 & 19.270 \\
\hline
2 & 494.6 & 24.972 & 3.313 & 22.794 & 24.817 & 27.642 \\
\hline
3 & 699.8 & 35.427 & 5.462 & 30.867 & 35.361 & 39.371 \\
\hline
4 & 904.8 & 40.056 & 7.220 & 36.594 & 40.675 & 43.795 \\
\hline
\end{tabular}
\label{tab2}
\end{center}
\end{table}

Being constructed on the \textit{2D representation} concept brings the ability to embed the solution into current differentiable rendering techniques, and thus helps reduce efforts in developing ideas in this paper for other research and applications. Although ideas need to be improved, I hope these contributions can motivate and benefit others' works. 

\section*{Acknowledgment}
I thank the Word (Genesis 1:3, John 1:1, Exodus 3:14), the Lord Jesus Christ, he is the actual owner of my work.

\appendices
\thispagestyle{plain}
\pagestyle{empty}
\section{Details about Circles in the Sky Dataset}
Let $E'$ is an $128 \times 256$ matrix of zero values. For each active point $E_{uv}$ in the \textit{2D representation} $E$, consider points in $E'$ have positions in $S=\{(u_{i}, v_{i}) \in \mathbb{N}^{2}|\exp(-\frac{\sqrt{(u_{i}-u)^2+(v_{i}-v)^2}}{3}) \ge 0.25, 1 \le u_{i} \le 128, 1 \le v_{i} \le 256\}^{N'}_{i=1}$, where $N'$ is the number of elements of $S$. A circles on $E'$ is drawn by following these steps for each $(u',v')$ in $S$: 
\begin{itemize}
\item If $E'_{u'v'}=0$, assign the value of $E_{uv}$ to $E'_{u'v'}$. 
\item Else, if  $E'_{u'v'} > E_{uv}$, assign the value of $E_{uv}$ to $E'_{u'v'}$. 
\end{itemize}

\section{Proof for Comment 1}
According to \eqref{eq:wL}, we consider two cases. For the first case, consider every screen position $(u',v')$ on $\widetilde{L}$ that have $q(\widetilde{I},(u',v')) \ne \emptyset$. Recall that in \eqref{eq:q}, 
\begin{equation}
\begin{split}
q(\widetilde{I},(u',v'))=&\{(rinv(\widetilde{I}_{ij}),\widetilde{I}_{ij})|\widetilde{I}_{ij} \ne 0, \nonumber \\
&(u',v')=\lfloor P(rinv(\widetilde{I}_{ij})) \rfloor \}_{i,j}. 
\end{split}
\end{equation}
Thus, for each $\widetilde{I}_{uv}$ satisfies the conditions in \eqref{eq:q}, we consider the corresponding kernel of this active point, $D_{uv}$. Let $({\Delta' u}, {\Delta' v})=(u'-{u}+c_{h},v'-{v}+c_{w})$, we have $\lim\limits_{c \to \infty} D_{uv{\Delta' u}{\Delta' v}}=1$. Indeed, because $\widetilde{I}_{uv} \ne 0$ and according to \eqref{eq:D}, we have $D_{uv{\Delta' u}{\Delta' v}}=\frac{\exp(c \norm{rinv(\widetilde{I}_{uv})-T_{uv{\Delta' u}{\Delta' v}}})}{\sum\limits_{i,j}\exp(c \norm{rinv(\widetilde{I}_{uv})-T_{uvij}})}$. And, this means that $\lim\limits_{c \to \infty} D_{uv{\Delta' u}{\Delta' v}}=1$ if the 3D point represented by $\widetilde{I}_{uv}$ is nearer to the \textit{XY-adjustment} $T_{uv{\Delta' u}{\Delta' v}}$ than other \textit{XY-adjustments} of $\widetilde{I}_{uv}$. Because we are considering points have the same depth value as indicated in \eqref{eq:Tz}, let's consider them in a Euclidean plane $\Pi:z=zinv(\widetilde{I}_{uv})$. Consider the set of \textit{XY-adjustments} of $\widetilde{I}_{uv}$, $\{T_{uvij}\}_{1 \le i \le c_{h},1 \le j \le c_{w}}$ in the plane $\Pi$, we have the set of points $V_{ij}=\{(x,y,z)|z=zinv(\widetilde{I}_{uv}),\lfloor P((x,y,z)) \rfloor=\lfloor P(T_{uvij}) \rfloor\}$ in the plane is the Voronoi cell associated with $T_{uvij}$ (based on the nature of $\lfloor \cdot \rceil$ function and its relationship with $\lfloor \cdot \rfloor$ function). According to \eqref{eq:q}, we have $(u',v')=\lfloor P(rinv(\widetilde{I}_{uv}) \rfloor$ and thus, according to \eqref{eq:TP}, the 3D point represented by $\widetilde{I}_{uv}$ is in the Voronoi cell of $T_{uv{\Delta' u}{\Delta' v}}$. Because points in the Voronoi cell associated with $T_{uvij}$ is nearer to the \textit{XY-adjustment} $T_{uvij}$ than other \textit{XY-adjustments}, we have the 3D point represented by $\widetilde{I}_{uv}$ is nearer to the \textit{XY-adjustment} $T_{uv{\Delta' u}{\Delta' v}}$ than other \textit{XY-adjustments} of $\widetilde{I}_{uv}$. Therefore, as said above, we have $\lim\limits_{c \to \infty} D_{uv{\Delta' u}{\Delta' v}}=1$. Let  $(r,s)=(u'-i+c_{h},v'-j+c_{w})$ and recall that $B_{ijkl}=D_{ijkl} \cdot \widetilde{I}_{ij}$, so, according to \eqref{eq:C}, we can rewrite \eqref{eq:q} as
\begin{equation}
\begin{split}
q(\widetilde{I},(u',v'))=&\{(rinv(\lim\limits_{c \to \infty} B_{rsij}), \lim\limits_{c \to \infty} B_{rsij})|\widetilde{I}_{ij} \ne 0, \\
&\lim\limits_{c \to \infty} D_{rsij}=1\}_{1 \le i \le c_{h}, 1 \le j \le c_{w}}. \\
=&\{(rinv(\lim\limits_{c \to \infty} B_{rsij}), \lim\limits_{c \to \infty} B_{rsij})| \\
&\lim\limits_{c \to \infty} B_{rsij} \ne 0\}_{1 \le i \le c_{h}, 1 \le j \le c_{w}}. 
\end{split}
\end{equation}

Consider the second case, screen positions $(u',v')$ on $\widetilde{L}$ that have $q(\widetilde{I},(u',v')) = \emptyset$. Thus, we have $B_{rsij} = 0$. 

In general, 
\begin{equation}
\widetilde{L}(u',v')=\left\{\begin{IEEEeqnarraybox}[\relax][c]{l's}
		rmin(\{(rinv(\lim\limits_{c \to \infty} B_{rsij}), \\
\lim\limits_{c \to \infty} B_{rsij)}\}_{k,l}), & if $\lim\limits_{c \to \infty} B_{rsij} \ne 0$, \\
		0, & otherwise. %
	\end{IEEEeqnarraybox}\right. 
\end{equation}
Finally, based on the definition of the function $rmin$, we have $\lim\limits_{c \to \infty} C_{u'v'} = \widetilde{L}^{128 \times 256}_{\widetilde{F},P,rmin}(u',v')$. 

\section{Proof for Comment 2}
Recall that the max value of the kernel $D_{{u}_{0} {v}_{0}}$ is $D_{{u}_{0} {v}_{0} \widetilde{k}_{0} \widetilde{l}_{0}} \approx 1$, while its other values are $0$. On the other hand, we have the expression $\abs{rinv(\lim\limits_{c \to \infty} \widetilde{L}_{ij})-rinv(L^{*}_{ij})}$ has a possible minimum value is $0$. Assume that $\abs{rinv(\lim\limits_{c \to \infty} \widetilde{L}_{ij})-rinv(L^{*}_{ij})}=0$, we have $rinv(\lim\limits_{c \to \infty} \widetilde{L}_{u'_{0} v'_{0}})=rinv(L^{*}_{u'_{0}v'_{0}})$. Thus, according to the definition of $\widetilde{L}$ in \textit{Comment~\ref{com:1}}, the definition of $L^{*}$ in \eqref{eq:Lstar}, and the assumption of a single point, these equations must be respectively satisfied: 
\begin{IEEEeqnarray}{c}
\widetilde{L}_{u'_{0} v'_{0}} = D_{(u'_{0}-\widetilde{k}_{0}+c_{h})(v'_{0}-\widetilde{l}_{0}+c_{w}) \widetilde{k}_{0} \widetilde{l}_{0}} \cdot \widetilde{I}_{(u'_{0}-\widetilde{k}_{0}+c_{h})(v'_{0}-\widetilde{l}_{0}+c_{w})}, \nonumber \\
\lim\limits_{c \to \infty} rinv(\widetilde{L}_{u'_{0} v'_{0}})=T_{{u}_{0} {v}_{0} (u'_{0}-{u}_{0}+c_{h}) (v'_{0}-{v}_{0}+c_{w})}, \nonumber \\
(u'_{0}-\widetilde{k}_{0}+c_{h},v'_{0}-\widetilde{l}_{0}+c_{w}) = ({u}_{0},{v}_{0}). \nonumber
\end{IEEEeqnarray}
The equation system is solved if $rinv(\widetilde{I}_{{u}_{0} {v}_{0}})=T_{{u}_{0} {v}_{0} (u'_{0}-{u}_{0}+c_{h}) (v'_{0}-{v}_{0}+c_{w})}$. Inversely, if the equation system is solved, we also have $\abs{rinv(\lim\limits_{c \to \infty} \widetilde{L}_{ij})-rinv(L^{*}_{ij})}=0$. Thus, based on the definition of $\theta_{opt}$, we have that if $\widetilde{\theta} = \theta_{opt}$ then $\mathbb{1}_{\mathbb{R}^{+}}(\widetilde{L})=L$. 

\section{Proof for Comment 3}
\begin{figure}[htbp]
\centerline{\includegraphics{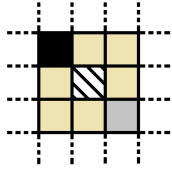}}
\caption{Illustration of the proof for Comment 3 on the projection plane. Brown squares represent squares have position in $\Upsilon'_{i}$. The black square represents the active point has position $(r,s) \in \Upsilon'_{i}$. The gray square represent $(r',s')$, the reflection of $(r,s)$ in $(u'^{0}_{i},v'^{0}_{i})$. }
\label{fig5}
\end{figure}

Consider an active point in $\widetilde{I}_{u_{i}v_{i}}$, where $(u_{i},v_{i})$ is in the set of active points in $\widetilde{I}$ ($W=\{(u_{i},v_{i})|\widetilde{I}_{u_{i} v_{i}} \ne 0\}^{M'}_{i=1}$), and its corresponding active point $\widetilde{L}_{u'_{i} v'_{i}}$ in $\widetilde{L}$, where $(u'^{0}_{i},v'^{0}_{i})$ is the \textit{proper screen position} of $\widetilde{I}_{u_{i}v_{i}}$. Let $(r,s)$ is a screen position in $\Upsilon'_{i}$ as illustrated in Fig.~\ref{fig5}. So, $(r,s)$ corresponds to a term of the first term in \eqref{eq:Ji}. Based on the assumptions of the \textit{Comment~\ref{com:3}}, we have that the distribution of the value of the first term is independent of $(u'^{0}_{i}, v'^{0}_{i})$ and, obviously, the first terms are independent random variables. 

Because of symmetry and the independent distribution of active points in $L$, we have that the absolute value of the expectation of the gradient of active points at the position $(r,s)$ is equal to the absolute value of the expectation of the gradient of active points at the reflection of $(r',s')$ in $(u'_{i},v'_{i})$. 
Moreover, we have gradients at these two screen position represent for two opposite transition directions of the 3D point represented by $\widetilde{I}_{u_{i}v_{i}}$. Thus sum of the expectation of the gradient of active points at $(r,s)$ and $(r',s')$ is zero. Apply the same fact for others points in $\Upsilon'_{i}$ we have the expectation of the first term is zero. 

From \eqref{eq:Je}, we have that the absolute values of terms in \eqref{eq:Ji} are less than a constant. Thus, along with the fact that the expectation of the first term is zero, we have its variance is less than a constant. 

\section{Proof for Lemma 1}
First, we will prove that if $(u,v)$ is the projection of $\widetilde{p}$ on $\widetilde{L}$ then it is also the projection of $p'$ on $L'$. According to \eqref{eq:Pr}, we have that if $(u,v)$ is the projection of $\widetilde{p}$ on $\widetilde{L}$ then $(u,v)$ must also be the projection of a point $q' \in Q'$. Assume that $q'$ is not $p'$. Let $A$ is the event of $(u,v)$ is the projection of a point in $Q'$ and $B$ is the event of $(u,v)$ is the projection of a point in $\widetilde{Q}$, we have $P(A)=1$ as $q'$ is deterministic and $P(B)=c$ (the assumption of the lemma). Because $P(A)=1$, we have the event $A \cap B$ is a necessary condition of the event of $\widetilde{L}$ has the same binary mask as $\widetilde{L} \odot L'$. So, we have 
\begin{IEEEeqnarray}{rCl}
P(\mathbb{1}_{\mathbb{R}^{+}}(L')=\mathbb{1}_{\mathbb{R}^{+}}(\widetilde{L} \odot L'))&<&P(A \cap B), \nonumber \\
&<&P(B), \nonumber \\
&<&c. 
\end{IEEEeqnarray}
Because $c \in (0,1)$, we have $P(\mathbb{1}_{\mathbb{R}^{+}}(L')=\mathbb{1}_{\mathbb{R}^{+}}(\widetilde{L} \odot L')) < 1$ (contrary to the assumption of the lemma in \eqref{eq:Pr}). So, $q'$ must be the same point as $p'$, this leads to $(u,v)$ is also the projection of $p'$ on $L'$. 

Without loss of generality, let $p=(x,y,z)$, $p'=(x + a,y + b,z)$ and $\widetilde{p}=(x + \widetilde{a},y + \widetilde{b},z)$. Because $(u,v)=\lfloor P(p') \rfloor=\lfloor P(\widetilde{p}) \rfloor$, we have 
\begin{equation}
|P(p')-P(\widetilde{p})| < C,
\label{eq:prf_c}
\end{equation}
where $C$ is a constant vector. According to the definition of linear perspective projection (without extrinsic parameters), we have 
\begin{equation}
|P(p')-P(\widetilde{p})| = \frac{(A p' - A \widetilde{p})}{z}, 
\label{eq:prf_z}
\end{equation}
where $A$ is a constant matrices. From \eqref{eq:prf_c} and \eqref{eq:prf_z}, we have $A \frac{p' - \widetilde{p}}{z} < C$. Along with the fact that $A$ is constant matrices, we know that there exist a constant vector $C'$, so that $\lim\limits_{z \to 0} (p' - \widetilde{p}) = \lim\limits_{z \to 0} C'z = (0,0,0)$. Because $p' - \widetilde{p}=(a-\widetilde{a}, b-\widetilde{b},0)$ and $\theta-\widetilde{\theta}=(a-\widetilde{a}, b-\widetilde{b})$, we have $\lim\limits_{z \to 0} (\theta-\widetilde{\theta}) = (0,0)$. Therefore, $\lim\limits_{z \to 0} \widetilde{\theta}=\theta$. 

\section{Pseudo-code of the Spatial Domain Mapping algorithm}
\makeatletter
\newcommand\fs@norules{\def\@fs@cfont{\bfseries}\let\@fs@capt\floatc@ruled
  \def\@fs@pre{}%
  \def\@fs@post{}%
  \def\@fs@mid{\kern3pt}%
  \let\@fs@iftopcapt\iftrue}
\makeatother
\floatstyle{norules}
\restylefloat{algorithm}
\begin{algorithm}[H]
\caption{Spatial Domain Mapping algorithm}\label{sdm}
\begin{algorithmic}[1]
\Require Computational graph of \textit{2D representation} $\widetilde{I}$, target binary mask $L$, original size of \textit{2D representation} $(H, W)=(128,256)$, order of kernel size $s \in \mathbb{N}^{+}$, zoom-out index $0 \le zoom \le s$
\Ensure Computational graph of \textit{proper 2D representation} $\widetilde{L}$
\Procedure{reshape}{$As$}\Comment{Reshape algorithm to reduce size of \textit{2D representations} by half in all spatial dimensions}
\State Create an empty list $B$
\For {$A \in As$}
\State Use row-major order \textit{reshape} method of Pytorch to reshape $A$ to shape $(h/2) \times 2 \times (w/2) \times 2$
\State Use row-major order \textit{permute} method of Pytorch to transpose $A$ to shape $4 \times (h/2) \times (w/2)$ while keeping the first and the second dimensions
\State Decouple the first order of $A$ to have a list of 4 new \textit{2D representations}
\State Store the list to $B$
\EndFor
\State \textbf{return} $B$
\EndProcedure
\Procedure{SDM3X3}{$Is$}\Comment{The SDM algorithm used $3 \times 3$ kernels}
\State Set an empty list $Ls$
\For {$I \in Is$}
\State Construct \textit{XY-adjustments} of active points of $I$
\State Appy the result of \textit{Comment~\ref{com:1}} to generate computational graph of \textit{proper 2D representation} $L'$ from $I$
\State Modify the computational graph of the \textit{proper 2D representation} to reflect \eqref{eq:Ji}
\State Store $L'$ to $Ls$
\EndFor
\State \textbf{return} $Ls$
\EndProcedure
\For{$i \gets 1,\, s$}
\State Construct a list of size-reduced \textit{2D representation}, 
\Statex $\widetilde{I} \gets reshape(\widetilde{I})$
\State $L \gets reshape(L)$
\EndFor
\State Perform SDM algorithm used $3 \times 3$ kernels, 
\Statex $\widetilde{L} \gets SDM3X3(\widetilde{I})$
\For{$i \gets 1,\, s-zoom$}
\State $\widetilde{I} \gets \widetilde{L}$
\State Increase (doubled in all dimensions) the size of \textit{2D representations} $\widetilde{I}$ and $L$ by inverting the reshape algorithm
\State $\widetilde{L} \gets SDM3X3(\widetilde{I})$
\EndFor
\State Cancel the gradient flow at screen positions of $\widetilde{I}$ which satisfy \eqref{eq:ignore}
\State Cancel the gradient flow at unactive screen positions of $\widetilde{L}$ as mentioned at \eqref{eq:grad_flow}
\State Construct zoomed-out target \textit{2D representation}, 
\Statex $L \gets \mathbb{1}_{\mathbb{R}^{+}}(\sum_{i}{L_i})$
\State \textbf{return} $\widetilde{L}$\Comment{\textit{proper 2D representation} $\widetilde{L}$ from $\widetilde{I}$}
\end{algorithmic}
\end{algorithm}

\end{document}